\documentclass[letterpaper, 10 pt, conference]{ieeeconf}  

\IEEEoverridecommandlockouts                              

\usepackage{graphicx} 
\usepackage{subfigure}
\usepackage[tbtags]{amsmath} 
\usepackage{amssymb}  
\usepackage{todonotes}
\usepackage{gensymb}
\usepackage[ruled,vlined,linesnumbered]{algorithm2e}
\usepackage{multicol}
\usepackage{multirow}
\usepackage{todonotes}
\usepackage{hhline}

\title{\LARGE \bf
Sample-efficient Deep Reinforcement Learning\\ with Imaginary Rollouts
 for Human-Robot Interaction
}

\author{Mohammad Thabet$^{1}$, Massimiliano Patacchiola$^{2}$, and Angelo Cangelosi$^{1, 3}$
\thanks{  $^{1}$Mohammad~Thabet~and~Angelo~Cangelosi~are~with~the~School~of Computer~Science,~the~University~of~Manchester,~United~Kingdom, \hspace{100pt} {\tt\small mohammad.thabet@manchester.ac.uk, angelo.cangelosi@manchester.ac.uk}}%
\thanks{$^{2}$Massimiliano Patacchiola is with the School of Informatics, the University of Edinbrugh, United Kingdom,
        {\tt\small mpatacch@ed.ac.uk}}%
\thanks{$^{3}$AIST Artificial Intelligence Research Centre, Tokyo, Japan.}
}

\begin{document}

\maketitle
\thispagestyle{empty}
\pagestyle{empty}

\begin{abstract}

Deep reinforcement learning has proven to be a great success in allowing agents to learn complex tasks. However, its application to actual robots can be prohibitively expensive. Furthermore, the unpredictability of human behavior in human-robot interaction tasks can hinder convergence to a good policy. In this paper, we present an architecture that allows agents to learn models of stochastic environments and use them to accelerate learning. We descirbe how an environment model can be learned online and used to generate synthetic transitions, as well as how an agent can leverage these synthetic data to accelerate learning. We validate our approach using an experiment in which a robotic arm has to complete a task composed of a series of actions based on human gestures. Results show that our approach leads to significantly faster learning, requiring much less interaction with the environment. Furthermore, we demonstrate how learned models can be used by a robot to produce optimal plans in real world applications.

\end{abstract}

\section{Introduction}

Deep reinforcement learning (RL) has been successfully applied to a variety of problems recently such as playing Atari games with super-human proficiency \cite{mnih2013playing}, and for robot control \cite{levine2016end}. However, Applying RL methods to real robots can be extremely costly, since acquiring thousands of episodes of interactions with the environment often requires a lot of time, and can lead to physical damage. Furthermore, in human-robot interaction (HRI) scenarios, human actions cannot be predicted with certainty, which can significantly impede convergence to a good policy.

One way of alleviating these problems is to have the agent learn a model of the environment, and use this model to generate synthetic data that can be used in conjunction with real data to train the agent. This assumes that the environment dynamics are easier to learn than an optimal policy, which is a generally valid assumption at least for some classes of tasks. Furthermore, if such a model is stochastic in nature, then the uncertainty in state changes can be taken into account, thus allowing more natural interaction with humans. Much like how people learn, an agent with a model of its environment can generate imaginary scenarios that can be used to help optimize its performance. This approach has garnered much attention in the field recently, and is sometimes refered to as endowing agents with \emph{imagination} \cite{racaniere2017imagination,kalweit2017uncertainty,ha2018recurrent}.

\begin{figure}[t]
  \centering
	  \includegraphics[width=0.4\textwidth]{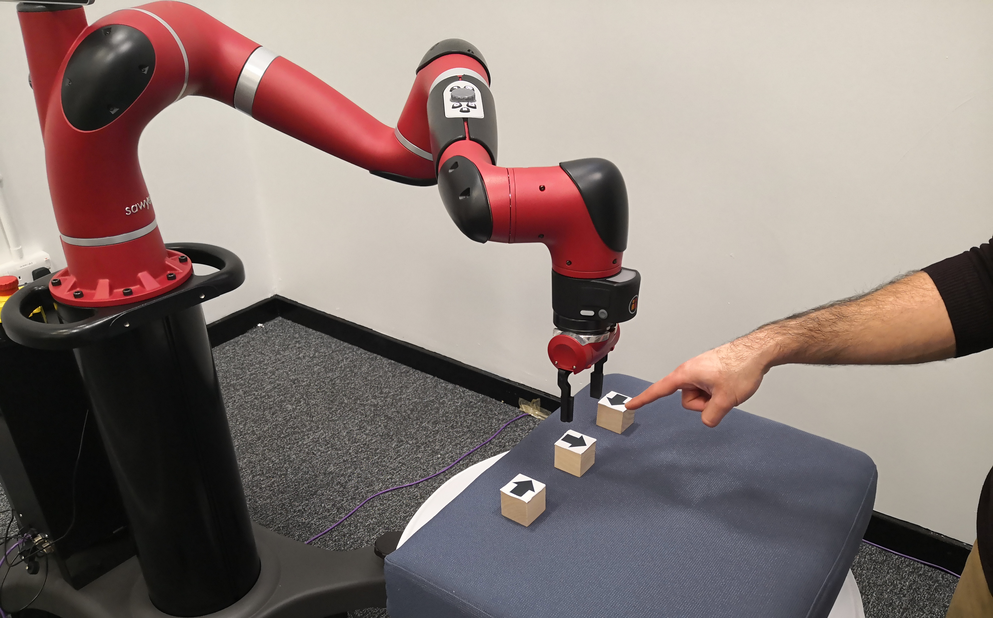}%
	  \label{fig:ro4}
  \caption[]{Experiments with the Sawyer robotic arm. The robot has to solve a puzzle by rotating the cubes to reach a goal state based on the pointing gesture by the human.}
  \label{fig:robot}
\end{figure}

In this paper we describe an architecture that allows an agent to learn a stochastic model of the environment and use it to accelerate learning in RL problems. In our approach, an agent encodes sensory information into low-dimensional representations, and learns a model of its environment online in latent space, while simultaneously learning an optimal policy. The model can be learned much faster than the policy, and therefore can be used to augment transitions collected from the real environment with synthetic transitions, improving the sample-efficiency of RL. Our approach requires no prior knowledge of the task; only the encoder needs to be pretrained on task-relevant images, and can generally be reused for multiple tasks. We test our architecture on a high-level robotic task in which a robot has to interpret a gesture and solve a puzzle based on it (Fig~\ref{fig:robot}). Results show that incorporating synthetic data leads to a significant speedup in learning, especially when only a small amount of real interaction data are made available to the agent.

\section{Related Work}

There has been much recent interest in the literature about combining model-free and model-based approaches to reinforcement learning. Ha and Schmidhuber \cite{ha2018recurrent} built models for various video game environments using a combination of a mixture density network (MDN) and a long short-term memory (LSTM), which they call MDN-RNN. In their approach, they first compress visual data into a low-dimensional representations via a variational autoencoder (VAE), and then train the MDN-RNN to predict future state vectors, which are used by the controller as additional information to select optimal actions. However, they pretrained the environment models on data collected by a random agent playing video games, whereas in our work a model for an HRI task is learned online.

The use of learned models to create synthetic training data has also been explored. Kalweit et al. \cite{kalweit2017uncertainty} used learned models to create imaginary rollouts to be used in conjunction with real rollouts. In their approach, they limit model usage based on an uncertainty estimate in the Q-function approximator, which they obtain with bootstrapping. They were able to achieve significantly faster learning on simulated continuous robot control tasks. However, they relied on well-defined, low-dimensional state representations such as joint states and velocities, as opposed to raw visual data as in our approach.

Racaniere et al. \cite{racaniere2017imagination} used a learned model to generate multiple imaginary rollouts, which they compress and aggregate to provide context for a controller that they train on classic video games. The advantage of this approach is that the controller can leverage important information contained in subsequences of imagined rollouts, and is more robust to erroneous model predictions.

Model rollouts can be used to improve targets for temporal differencing (TD) algortihms as well. Feinburg et al. \cite{feinberg2018model} used a model rollout to compute improved targets over many steps, in what they termed model-based value expansion (MVE). More recently, Buckman et al. \cite{buckman2018sample} propsed an extension to MVE, in which they use an ensemble of models to generate multiple rollouts of various lengths, interpolating between them and favoring those with lower uncertainty.

Deep reinforcement learning is increasingly being employed successfully for robots in continuous control tasks and manipulation \cite{levine2016end,lillicrap2015continuous,gu2017deep,popov2017data}. However, its application to high-level tasks and in HRI has been very limited. Qureshi et al. \cite{qureshi2016robot} used a multimodal DQN to teach a humanoid robot basic social skills such as successfully shaking hands with strangers. Input to their system consisted of depth and greyscale images. Interaction data were collected using the robot over a period of 14 days, where they have separated the data collection and training phases and alternated between them for practical reasons. In our work, we are primarily interested in increasing the sample efficiency so that training requires less resources, allowing RL to be more practical for robots.

\section{Background}
\subsection{Reinforcement Learning}
In reinfrocement learning \cite{sutton2018reinforcement}, a task is modelled as a Markov decision process (MDP) where an agent influences its environment state $s_t$ with action $a_t = \pi(s_t)$ chosen according to some policy $\pi$. The environment then transitions into a new state $s_{t+1}$ and provides the agent with a reward signal $r_{t}$. This process repeats until the environment reaches a terminal state, concluding an episode of interaction. The goal of the agent is to learn an optimal policy $\pi^*$ and use it to maximize the expected return, which is the discounted sum of rewards, $R_t = \sum_{t=0}^{T} \gamma^t r_t$ where $\gamma \in [0,1]$ is the discount factor and T is the timestep a terminal state is reached.

There are model-based and model-free algorithms to find an optimal policy. One such model-free method is to learn the action-value function $Q^\pi(s, a)$, which is the expected return from taking action $a$ in state $s$ and following policy $\pi$ thereafter: $Q^\pi(s_t, a_t) = \mathbb{E}_\pi[R_t|s_t, a_t]$. The agent's goal thus becomes to learn an optimal Q-function $Q^*(s, a) = \mathrm{max}_\pi Q^\pi (s, a)$. This can be achieved using a recursive relationship known as the \emph{Bellman equation}:
\begin{equation}
Q^*(s_t,a_t) = \mathbb{E} [r_t + \gamma \mathop{\mathrm{max}}_{a_{t+1}} Q^*(s_{t+1}, a_{t+1}) | s_t, a_t]
\label{eq:bellman}
\end{equation}
Since most non-trivial tasks have very large state or action spaces, usually the Q-function cannot be calculated analytically, and is estimated instead by a function approximator $Q_\theta(s,a)$ with parameters $\theta$. One common approach is deep Q-networks (DQN) \cite{mnih2013playing}, in which transitions are stored as tuples of the form $(s_t, a_t, r_t, s_{t+1})$, and used to train a neural network so that $Q_\theta(s,a) \approx Q^*(s,a)$. A DQN is trained to minimize the loss function:
\begin{equation}
\mathcal{L}(\theta) = (y_t - Q_\theta (s_t, a_t))^2
\label{eq:DQN}
\end{equation} 
where the target $y_t$ is obtained from Equation~\ref{eq:bellman} using the estimate $Q_\theta (s_{t+1}, a_{t+1})$. Given the gradients of Equation~\ref{eq:DQN} with respect to $\theta$, the network can be trained using some variation of stochastic gradient descent. Actions are chosen based on the $\epsilon$-greedy policy where the optimal action is chosen with probability $1-\epsilon$ and a random action with probability $\epsilon$.

\subsection{Variational Autoencoders}

Variational autoencoders (VAE) \cite{kingma2013auto} are generative models that can be used to both generate synthetic data, and to encode existing data into representations in low-dimensional latent space. Like traditional autoencoders, they consist of an encoding network and a decoding one. The key difference is that they encode a data point $x$ into a probability distribution over latent variable vector $z$. The goal is then to learn an approximate multivariate Gaussian posterior distribution $q(z|x) = \mathcal{N}(\mathbf{\mu}(x), \Sigma(x)\mathbf{I})$ which is assumed to have a unit Gaussian prior $p(z) = \mathcal{N}(\mathbf{0},\mathbf{I})$. This can be done by minimizing the Kullback-Liebler (KL) divergence between $q(z|x)$ and the true posterior $p(z|x)$:
\begin{equation}
\begin{aligned}[b]
& \mathrm{KL}(q(z|x)\,||\,p(z|x))\\
& = \mathbb{E}_q [\log\,q(z|x) - \log\,p(z|x)]	\\
& = \mathbb{E}_q [\log\,q(z|x) - \log\,p(x|z) - \log\,p(z) - \log\,p(x)]	\\
& = -\mathbb{E}_q [\log\,p(x|z)] + \mathrm{KL}(q(z|x)\,||\,p(z)) - \log\,p(x)
\end{aligned}
\label{eq:vae_kl}
\end{equation}
Here, the first term is the reconstruction loss, while the second term penalizes divergence of the learned posterior from the assumed proir. The expectation $\mathbb{E}_q [log\,p(x|z)]$ can be approximated as $log\,p(x|z)$ by sampling a vector $z = \mathbf{\mu}(x) + \Sigma^{1/2}(x) \ast \epsilon$ with $\epsilon \sim \mathcal{N}(\mathbf{0},\mathbf{I})$ and decoding it with the decoder network. Since maximizing the marginal likelihood $p(x)$ is also maximizing the expected likelihood $\mathbb{E}_q [log\,p(x|z)]$, and since $p(z) = \mathcal{N}(\mathbf{0},\mathbf{I})$, minimizing Equation~\ref{eq:vae_kl} is equivalent to minimzing:
\begin{equation}
\mathcal{L(\theta, \phi)} = -\log\,p_\phi(x|z) + \frac{1}{2}\sum_{j=1}^{J} (1 + log\,\sigma_j^2 - \mu_j^2 - \sigma_j^2),
\label{eq:vae_loss}
\end{equation}
where we have parametrized the encoder and decoder networks with $\theta$ and $\phi$ respectively, $J$ is the dimensionality of the latent space, and $\sigma_j$ are the diagonal elements of $\Sigma(x;\theta)$. The encoder and decoder networks are trained back to back to minimize the loss given by Equation~\ref{eq:vae_loss}. Note that if $p(x|z)$ is Bernoulli, the reconstruction loss is equivalent to the cross-entropy between the actual $x$ and the predicted $\hat{x}$.

\subsection{Mixture Density Networks}

Mixture density networks (MDN) \cite{bishop1994mixture} are neural networks that model data as a mixture of Gaussian distributions. This is useful for modeling multi-valued mappings, such as many inverse functions or stochastic processes. MDNs model the distribution of target data $y$ conditioned on input data $x$ as $p(y|x) = \sum_{i=1}^{m} \alpha_i(x)\, \phi(y;\mu_i(x),\sigma_i(x)^2)$ where $m$ is the number of components, $\alpha_i$ are the mixture coefficients subject to $\sum_{i=1}^{m} \alpha_i = 1$, and $\phi(\cdot;\mu, \sigma^2)$ is a Gaussian kernel with mean $\mu$ and variance $\sigma^2$. MDNs have a similar structure to feedforward networks, except that they have three parallel output layers for three vectors: one for the means, one the variances, and one for the mixture coefficients. The network parameters $\theta$ are optimized by minimizing the negative log-likelihood of the data:
\begin{equation}
\mathcal{L(\theta)} = -\log\,\sum_{i=1}^{m} \alpha_i(x;\theta)\, \phi(y;\mu_i(x;\theta),\sigma_i(x;\theta)^2)
\end{equation}

To predict an ouput for a given input, we sample from the resulting mixture distribution by first sampling from categorical distribution defined by $\alpha_i$ to select a component Gaussian, and then sampling from the latter.

\section{Architecture}

\begin{figure}[t]
  \centering
	  \includegraphics[width=0.47\textwidth]{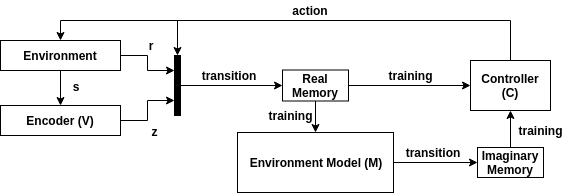}%
	  \label{fig:ro4}
  \caption[]{Overview of the proposed architecture. The controller C influences the environment with an action, which produces state \(s\) and reward \(r\). The encoder V encodes \(s\) into latent state vector \(z\). The environment model M can be trained on real transitions and then used to generate imaginary transitions. C can then be trained on both real and imaginary transitions.}
  \label{fig:architecture}
\end{figure}

The proposed architecture consists of three components: the vision module (V) that produces abstract representations of input images, the environment model (M) which generates imaginary rollouts, and the controller (C) that learns to map states into actions. We assume that the environment is Markovian and is fully represented at any given time by the input image. Figure \ref{fig:architecture} shows an overview of the architecture.

V comprises the encoder part of a variational auto-encoder (VAE) \cite{kingma2013auto}, and is responsible for mapping the high-dimensional input images into low-dimensional state representations. The controller and the environment model are trained in this low-dimensional latent space, which is generally computationally less expensive. The main advantage of using a VAE instead of a vanilla auto-encoder is that the VAE maps every input image into a continuous region in the latent space, defined by the parameters of a multivariate Gaussian distribution. This makes the environment model more robust and ensures that its output is meaningful and can be mapped back into realistic images. 

M is responsible for generating synthetic transitions, and predicts future states \(z_{t+1}\) and the reward \(r_t\) based on current states \(z_t\) and input actions \(a_t\). It is composed of three models: a mixture density network (MDN) \cite{bishop1994mixture} that learns the transition dynamics, a reward predictor called the r-network, and a terminal state predictor called the d-network. The MDN learns the conditional probability distribution of the next state \( p(z_{t+1}|z_t, a_t) \). The r-network learns to predict the reward for each state, while the d-network learns to predict whether a state is terminal or not. Both the r- and d-networks are implemented as feed-froward neural networks. To generate imaginary rollouts, M can be seeded with an initial state from V, and then run in closed loop where its output is fed back into its input along with the selected action. The advantage of using an MDN is that it is possible to learn a model of stochastic environments, in which an action taken in a given state can lead to multiple next states. This is especially useful for use in HRI tasks, in which the human response to actions taken by the robot cannot be expected with certainty. Furthermore, modelling the next state probabilistically is much more robust to errors in prediction, allowing the environment model to run in closed loop.

Lastly, C is responsible for selecting the appropriate action in a given state. It is implemented as a Q-network, and learns to estimate the action values for states. C is trained on both real and imaginary transitions to maximize the cumulative reward.

\section{Experiments}

The experiments detailed in this section are designed to evaluate our approach on a real world robotic application. We are interested primarily in the performance increase gained by utilizing the learned model, compared to the baseline DQN method \cite{mnih2013playing}.

\subsection{Experiment Setup}

To test our architecture, we desinged a task in which a robot has to solve a puzzle based on pointing gestures made by a human. The robot sees three cubes with arrows painted on them, with each arrow pointing either up, down, left, or right. The human can point to any of the three cubes, but may not point to a different cube during the same episode. To successfully complete the task, the robot has to rotate the cubes so that only the arrow on the cube being pointed to is in the direction of the human, with the constraint that at no point should two arrows point to the same direction. The task is similar to puzzle games typically used in studies about robot learning, such as the Towers of Hanoi puzzle \cite{lee2013syntactic}.

\begin{figure}[htbp]
  \centering
  \subfigure[][]{%
	  \includegraphics[width=0.12\textwidth]{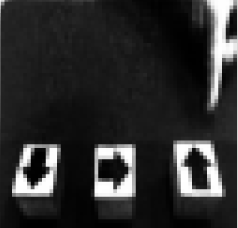}%
	  \label{fig:goal_state}
	  } \hspace{5 mm}
	\subfigure[][]{%
	  \includegraphics[width=0.12\textwidth]{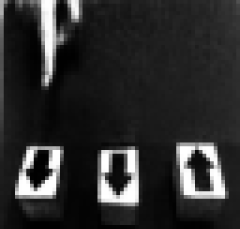}%
	  \label{fig:illegal_state}
	  }
  \caption[]{Examples of terminal states of the task. \subref{fig:goal_state} is a goal state, while \subref{fig:illegal_state} is an illegal state.}
  \label{fig:states}
\end{figure}

We formulate the task as an RL problem in which the agent can choose from 6 discrete actions at any given time: rotating any of the three cubes 90\degree clockwise or counterclockwise. The robot gets a reward of +50 for reaching the goal state, -5 for reaching an illegal state, and -1 otherwise to incentivize solving the task as efficiently as possible. An episode terminates if the robot reaches either a goal state or an illegal state, or after it performs 10 actions.  See Fig~\ref{fig:states} for examples of goal and illegal terminal states.

To train the robot, we created a simulated environment that receives the selected action from the agent (the robot) as input, and outputs an image representing its state, along with a reward signal and a flag to mark terminal states. The environment is implemented as a finite state machine with 192 macrostates, where each macrostate is the combination of 3 microstates describing the directions of each of the three arrows, plus another microstate describing which box the hand is pointing to. Whenever the environment transitions to a certain state, it outputs one of a multitude of images associated with that state at random, thus producing the observable state that the agent perceives.

To produce the images used for the environment, we first collected multiple image fragments for each of the possible microstates of the environment. Each of these fragments depicts a slight variation for the same microstate, for example slightly different box positions or hand positions. We thus create a pool of multiple image fragments for each possible microstate. To synthesize a complete image for a given macrostate, we choose a random fragment for each of its constituent microstates, and then patch them together. For the experiments, we collected 50 fragments for each possible hand microstate, and 16 fragments for each possible arrow microstate, resulting in about \(4 \times 10^7\) possible unique synthesized images. The images were taken with the Sawyer robotic arm camera (Fig.~\ref{fig:robot}). For the experiments, we synthesized 100,000 training images, and 10,000 test images.

\subsection{Procedure}

The training procedure for our experiments can be summarized as follows:

\begin{enumerate}
\item Train the VAE on all training images.
\item Start collecting real rollouts and training the controller.
\item After some amount of episodes, start training environment model M on collected rollouts.
\item Use M to generate synthetic rollouts simultaneously with real rollouts.
\item Continue training the controller using both real and synthetic rollouts.
\end{enumerate}

The exact training procedure is given in Algorithm~\ref{alg:alg1}. In the following, we will detail the training procedure for each component of the system and justify our choice of different parameters.

\begin{algorithm}[t]
\caption{Training procedure for agents.}
\label{alg:alg1}
\SetKwInput{KwRequire}{Require}
	\KwRequire{Pretrained encoder $V$ }
	Initialize controller $C$ and environment model $E_{\theta}$	\\
	Initialize real memory $M_R$ and imaginary memory $M_I$	\\
	\For{$e = 0$ \KwTo $num\_episodes$}{
		Observe initial state $s_0$	\\
		$s_t = s_0$	\\
		\While{$s_t$ is not terminal}{
			Use $V$ to encode $s_t$ into $\mu_t$ and $\sigma_t$	\\
			Apply action $a_t=C(\mu_t)$	\\
			Observe $s_{t+1}$, reward $r_t$, terminal signal $d_{t+1}$	\\
			Encode $s_{t+1}$ into $\mu_{t+1}$ and $\sigma_{t+1}$	\\
			Save transition $(\mu_t, \sigma_t, a_t, \mu_{t+1}, \sigma_{t+1}, r_t, d_{t+1})$ in $M_R$	\\
			\For{$i = 0$ \KwTo $N_E$}{
				Train $E_{\theta}$ on minibatch from $M_R$
			}
			\For{$i = 0$ \KwTo $N_R$}{
				Train $C$ on minibatch from $M_R$
			}
			\If{$e \geq I_{start}$}{
				Use $E_{\theta}$ to generate $I_B$ imaginary rollouts of depth $I_D$	\\
				Save imaginary transitions in $M_I$	\\
				\For{$i = 0$ \KwTo $N_I$}{
					Train $C$ on minibatch from $M_I$
				}
			}
			$s_t = s_{t+1}$
		}
	}
\end{algorithm}

\subsubsection{Variational Autoencoder}

To train the VAE, we split the grayscale training images along the horizontal axis into 3 strips, where each strip contains a single box. We then fed the strips into the VAE as 3 channels to help the VAE learn more task-relevant features. The architecture used for the VAE is that used by Ha and Schmidhuber in \cite{ha2018recurrent}, except that we encode the images into 8-dimentional latent space. The VAE was trained on the 100,000 synthesized training images after scaling them down to a manageable $64 \times 64$ resolution for 1000 epochs. The VAE is trained to minimize the combined reconstruction and KL losses given by Equation~\ref{eq:vae_loss}. Here, the reconstruction loss is given by the pixel-wise binary cross-entropy. The KL loss was multiplied by a weighting factor \(\beta\) that controls the capacity of the latent information chanel. In general, increasing \(\beta\) yields more efficient compression of the inputs and leads to learning independent and disentagled features, at the cost of worse reconstruction \cite{higgins2017beta}. We found $\beta=4$ to produce best results. The Adam optimizer was used with a learning rate of 0.0005 and a batch size of 2000.

\subsubsection{Environment Model}

The MDN used to model the dynamics in the environment model learns the posterior probability of the next latent state vector as a Gaussian mixture model with 5 components. The MDN has 3 hidden layers of 256 ReLU units with 50\% dropout and 3 parallel output layers for the distibution parameters: one for the mixture coefficients with softmax activation, one for means and one for variances both with linear activation. When collecting transitions, we stored the parameters \(\mu\) and \(\sigma\) produced by V for each frame, and we sampled from \(\mathcal{N}(\mu, \sigma)\) to obtain latent space vectors when constructing a training batch. This form of data augmentation was found to greatly improve the generalization and performance of the model. The accompanying r-network has 3 hidden layers of 512 ReLU units each with 50\% dropout, and was trained to minimize the logcosh loss. The d-network has 2 hidden layers of 256 ReLU units each with 50\% dropout and was trained to minimize the binary crossentropy. Both networks were trained to predict the corresponding value based on the state alone. During training, the MDN, the r-network and the d-network were all updated 16 times on batches of 512 transitions each timestep using the Adam optimizier with a learning rate of 0.001.

\subsubsection{Controller}

The controller is a DQN consisting of 3 hidden layers (512 ReLU, 256 ReLU, 128 ReLU) and a linear output layer. It was updated once on a batch of 64 real transitions and once on a batch of 64 imaginary transitions each timestep. We found that for such a relatively simple task, updating the controller more often led to worse performance. We also found that using popular DQN extensions like a separate target network or prioritized experience replay did not significantly affect performance. The controller used an \(\epsilon\)-greedy strategy with an exponentially annealing exploration rate given by \(\epsilon = \epsilon_{min} + (\epsilon_\text{max} - \epsilon_\text{min})e^{-\lambda t}\) with \(\epsilon_\text{min} = 0.001\), \(\epsilon_\text{max} = 0.8\), \(\lambda = 0.03\), and \(t\) is the time step. The controller was trained to minimize the MSE loss using an Adam optimizer with a learning rate of 0.001.

\subsubsection{Parameters}

When training the agents, we set the depth of imaginary rollouts $I_D$ 10, and the breadth $I_B$ to 3.
The size of the real memory was 50,000 transitions, and that of the imaginary memory was 3,000. We found that training the controller only on recently generated transitions leads to better performance, since more recent copies of $M$ produce better predictions. We achieve this by both limiting the imaginary memory size, and generating multiple rollouts simultaneously. Furthermore, we found that setting the update rate of the controller on both real and imaginary transsitions ($N_R$ and $N_I$ in Algorithm~\ref{alg:alg1}) to more than 1 can lead to stability issues. Another parameter we had to tune was the number of episodes to wait before staring to generate imaginary rollouts ($I_{start}$ in Algorithm~\ref{alg:alg1}), since M will produce erroneous predictions early on in the training. We found that waiting for about 1000 episodes provides best results.

\subsection{Results}

We compare the performance of agents augmented with imaginary transitions using our approach with a baseline DQN trained only on real transitions. To aid comparison, all hyperparameters and architectural choices were the same for agents augmented with our approach and the baseline DQN. For a given number of training episodes, we trained 5 agents from scratch and then tested them on the simulated environment for 1000 episodes. We then averaged the percentage of successfully completed episodes of all 5 agents in all test runs.

\begin{figure}[htbp]
  \centering
  \subfigure[][]{%
	  \includegraphics[width=0.23\textwidth]{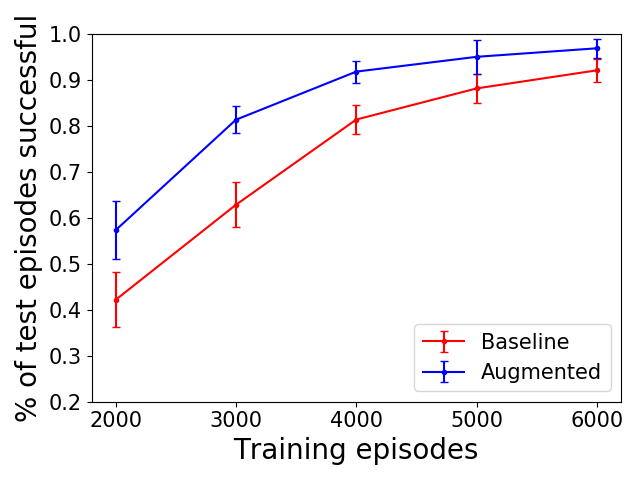}%
	  \label{fig:res_easy}
	  }
	\subfigure[][]{%
	  \includegraphics[width=0.23\textwidth]{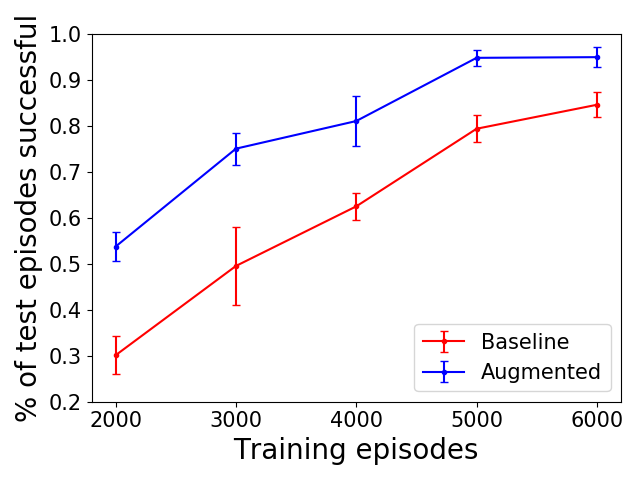}%
	  \label{fig:res_hard}
	  }
	 \subfigure[][]{%
	  \includegraphics[width=0.3\textwidth]{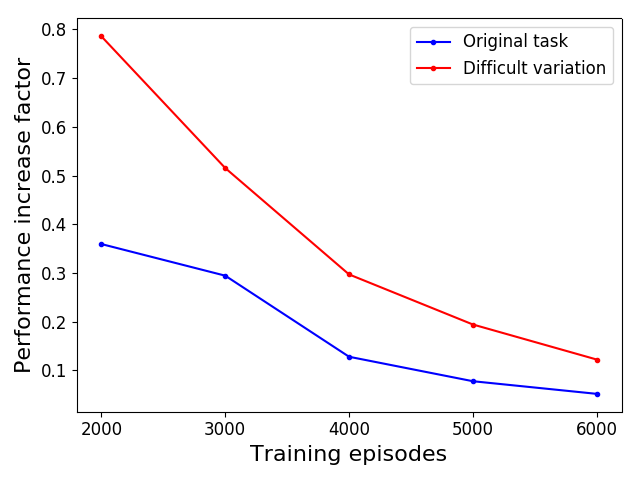}%
	  \label{fig:perf}
	  }
  \caption[]{Test results for various numbers of training episodes, \subref{fig:res_easy} for the original task, and \subref{fig:res_hard} for the more difficult variation of the task. \subref{fig:perf} shows the performance increase in both tasks. Error bars represent standard deviations.}
  \label{fig:results}
\end{figure}

The agents trained using our approach performed significantly better than baseline DQN when trained for a small amount of episodes, with an increase of 35.9\% in performance at 2000 episodes (Fig~\ref{fig:res_easy}). The advantage then starts to decline the more episodes the agent is trained, as the baseline DQN catches up quickly. This is to be expected since at higher episodes, the agent has collected enough real tranisitons and no longer needs the extra data generated by the environment model. Table~\ref{tab:easy} shows the exact results for this experiment.

We then increased the difficulty of the task while keeping the dynamics the same by additionally requiring the goal state not to have any arrows pointing towards the agent, and ran all the tests again. Results can be seen in Fig~\ref{fig:res_hard} and Table~\ref{tab:hard}. Augmented agents showed even greater performance increase compared to the baseline DQN, with up to 78.5\% increase in performance at 2000 training episodes. This shows that the performance increase due to using synthetic transitions is proportional to the difference in complexity between the task itself and the environment dynamics.

\begin{table}[h]
\centering
\begin{tabular}{|l||c|c|c|}
\hline
Episodes & Base DQN     & Augmented    & \% increase \\ \hline
2000     & 42.18 (6.01) & 57.34 (6.37) &   35.94     \\ \hline
3000     & 62.88 (4.91) & 81.4 (2.95)  &   29.45     \\ \hline
4000     & 81.44 (3.13) & 91.88 (2.38) &   12.81     \\ \hline
5000     & 88.22 (3.07) & 95.1 (3.76)  &    7.79     \\ \hline
6000     & 92.16 (2.57) & 96.96 (2.1)  &    5.2      \\ \hline
\end{tabular}
\caption{Mean percentage of successful test episodes for various numbers of training episodes for the original task. Std. deviations are given in parenthesis. For reference, a random agent scored 3.72\%.}
\label{tab:easy}
\end{table}

\begin{table}[h]
\centering
\begin{tabular}{|l||c|c|c|}
\hline
Episodes & Base DQN     & Augmented    & \% increase \\ \hline
2000     & 30.12 (4.16) & 53.78 (3.16) &   78.55     \\ \hline
3000     & 49.56 (8.48) & 75.1 (3.5)   &   51.53     \\ \hline
4000     & 62.54 (2.92) & 81.12 (5.35) &   29.7      \\ \hline
5000     & 79.44 (2.97) & 94.88 (1.8)  &   19.43     \\ \hline
6000     & 84.66 (2.73) & 95.03 (2.23) &   12.24      \\ \hline
\end{tabular}
\caption{Mean percentage of successful test episodes for various numbers of training episodes for the difficult variation of the task. Std. deviations are given in parenthesis. For reference, a random agent scored 3.39\%.}
\label{tab:hard}
\end{table}

\subsubsection*{Generating Plans}

\begin{figure}[htbp]
  \centering
  \subfigure[][]{%
	  \includegraphics[width=0.12\textwidth]{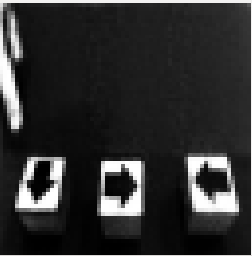}%
	  \label{fig:ro1}
	  } 
  \subfigure[][]{%
	  \includegraphics[width=0.12\textwidth]{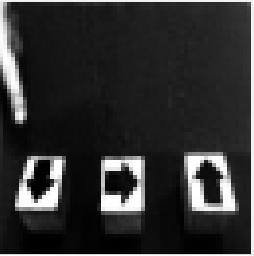}%
	  \label{fig:ro2}
	  }
  \subfigure[][]{%
	  \includegraphics[width=0.12\textwidth]{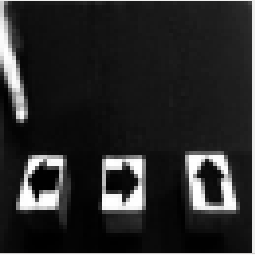}%
	  \label{fig:ro3}
	  }
  \subfigure[][]{%
	  \includegraphics[width=0.12\textwidth]{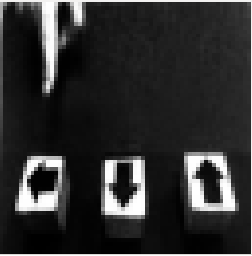}%
	  \label{fig:ro4}
	  }
  \subfigure[][]{%
	  \includegraphics[width=0.12\textwidth]{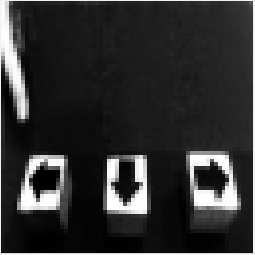}%
	  \label{fig:ro5}
	  }
  \subfigure[][]{%
	  \includegraphics[width=0.12\textwidth]{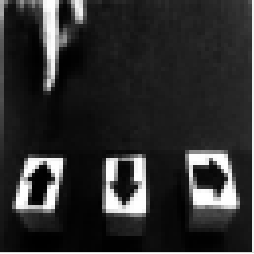}%
	  \label{fig:ro6}
	  }
  \caption[]{An example of an imaginary rollout of length 5. \subref{fig:ro1} is the initial state as observed by the robot. \subref{fig:ro2} through \subref{fig:ro6} are imagined next states after successively applying actions in the optimal plan. The visualizations of the model predictions were obtained by mapping the latent space vectors to images via the decoder part of the VAE.}
  \label{fig:im_ro}
\end{figure}

One of the advantages of learning an environment model is that it allows a trained agent to produce entire plans given only the initial state\footnote{This is only possible for environments with deterministic underlying dynamics}. This can be achieved by initializing the environment model with the initial state, and then generating an imaginary rollout in which the controller always chooses the optimal action for each state. To demonstrate this, we deployed a controller and an environment model on the Sawyer robotic arm (Fig~\ref{fig:robot}), where both networks had been previously trained for 6000 episodes using the training method descirbed previously. Afterwards, we ran experiments to evaluate the planning capabilities of the system. Each experiment began by setting the cubes to a random state, with the experimenter pointing to a random cube. Then, we let the robot observe the configuration with the camera, and asked it to produce a plan consisting of a trajectory of actions to solve the task in its original form. The robot can execute the plan by selecting successively from a set of pre-programmed point-to-point movements to rotate the boxes. Out of 20 test runs, the robot successfully solved the task 17 times. The correct generated plans varied in length from 1 to 5, depending on the initial state. Moreover, the generated plans for all successful runs were optimal, containing only the fewest possible actions required to solve the task. Fig~\ref{fig:im_ro} shows an example of an imaginary rollout according to an optimal plan of length 5 as generated by the agent.

\subsubsection*{Model Generalization}

\begin{figure}[htbp]
  \centering
  \subfigure[][]{%
	  \includegraphics[width=0.12\textwidth]{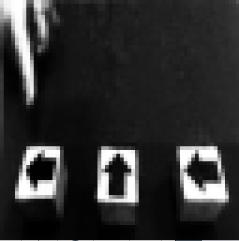}%
	  \label{fig:trans1}
	  } \hspace{5 mm}
	\subfigure[][]{%
	  \includegraphics[width=0.12\textwidth]{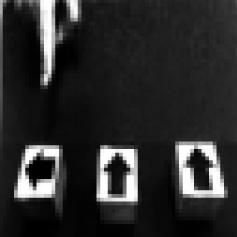}%
	  \label{fig:trans2}
	  }
  \caption[]{An example of model prediction for unseen transitions. The action selected here is to rotate the rightmost cube clockwise. \subref{fig:trans1} is the state before the action, and \subref{fig:trans2} is the state after.}
  \label{fig:unseen_trans}
\end{figure}

One of the interesting results we noticed is that the model showed some generalization capabilities to transitions it had not experienced before. Since episodes always terminated after encountering a terminal state, the model never experienced any transitions from this kind of state. To test model generalization, we deliberately set the model state to a random termminal state 20 times, and then asked it to predict the next state for a random action each time. A model trained for 5000 episodes was able to correctly predict the next state 75 \% of the time. Fig~\ref{fig:unseen_trans} shows an example of model prediction for unseen transitions.

\subsection{Discussion}

One of the main challenges in learning a model online is avoiding overfitting on the small subset of data that are made available early in the training. A model can easily get stuck in a local minimum if it gets trained execcsively on initial data, and fail to converge later to an acceptable loss value in a reasonable amount of time as more data are made available\footnote{When trained online, high-capacity models often exhibited a behaviour reminiscent of the Dunning-Kruger effect. They would achieve a very low loss value early in the training, which would quickly rise as more data are acquired, before eventually settling at a value in between.}. We achieve this through three things. First, we limit the model capacity by deliberately choosing smaller model sizes. Second, we adopt a probabilistic approach to encoding latent space representations and modeling environment dynamics. Third, we employ high dropout rates in the models. We also found that selecting an unnecessarily large latent space dimensionality leads to worse models.

Probabilistic models are also much more robust, which is essential when using the dynamics model in closed loop to generate rollouts. Traditional models based on point estimates will produce some error in prediction, which will quickly compound resulting in completely erroneous predictions sometimes as early as the second pass. This of course makes using imaginary rollouts detrimental to learning.

The ability to learn stochastic models can be useful even for environments whose underlying dynamics are deterministic. An environment with deterministic underlying dynamics can have stochastic observable dynamics, since each latent state of the environment can produce multiple observable states. For example, the task we used for the experiments has deterministic underlying dynamics, since the configuration of the arrows will alwyas change in the same way in response to a certain action. However, the observable state will change stochastically. The positions of the boxes or the hand may differ for the same configuration. The agent has no knowledge of the underlying dynamics since it only has access to observable states. Therefore, it needs to be able to model the observable dynamics stochastically in order to produces realistic imaginary rollouts. 

The generalization capabilities of the dynamics model can in principle be used to facilitate learning other similar tasks. The two variations of the task we used for the experiments share the exact same dynamics; they are only different in the definition of the reward functions. Indeed, for any given dynamics, an arbitrarily large family of tasks can be defined by specifying different reward functions. If learning the reward function can be separated from learning the dynamics, and assuming that the former is easier to learn than the latter, then learning new tasks in the same family will become much faster once the agent learns a dynamics model. However, this is left for future work. 

\section{Conclusion}

In this paper we presented an architecture that allows an agent to learn a model of stochastic environments in a reinforcement learning setting. This allows the agent to significantly reduce the amount of interactions it needs to make with the actual environment. This is especially useful for tasks involving real robots in which collecting real data can be prohibitively expensive. Furthermore, the ability to model stochastic environments makes this approach well-suited for tasks involivng interaction with humans as their actions usually cannot be predicted with certainty. We provided a detailed algorithm describing how to train both the agent and the environment model simultaneously, and how to use both synthetic data in conjunction with real data. We validated our approach on a high-level robotic task in which an agent has to simultaneously interpret a human gesture and solve a puzzle based on it. Results show that agents augmented with synthetic data outperform baseline methods especially in situations where only limited interaction data with the environment are available.

In future work, we will include recurrent models (such as LSTMs) in our architecture to handle environments with non-Markovian state representations. Furthermore, we will experiment with building environment models that can capture multi-modal dynamics, allowing agents to make use of acoustic information for instance. Another important extension is to include a measure of uncertainty to limit model usage if its output is erroneous. The simplest way to achieve this is by using model ensembles. We will also incorporate different ways of leveraging synthetic data to improve data efficiency even further, such as using imaginary rollouts to compute improved targets and for predicting future outcomes directly. Finally, we will investigate using programming by demonstration techniques \cite{billard2008robot} to bootstrap agents, further decreasing the amount of interactions the robot has to make with the environment.




\section*{Acknowledgment}

This project has received funding from the European Union's Horizon
2020 framework programme for research and innovation under the Marie
Sklodowska-Curie Grant Agreement No.642667 (SECURE)

\bibliographystyle{IEEEtran}
\bibliography{references}

\end{document}